%% file: main.tex
\documentclass{article}

\usepackage{PRIMEarxiv}

\usepackage[utf8]{inputenc} 
\usepackage[T1]{fontenc}    
\usepackage{hyperref}       
\usepackage{url}            
\usepackage{booktabs}       
\usepackage{amsfonts}       
\usepackage{nicefrac}       
\usepackage{microtype}      
\usepackage{lipsum}
\usepackage{fancyhdr}       
\usepackage{graphicx}       
\graphicspath{{media/}}     

\usepackage{tabularx}
\newcolumntype{k}{X}
\newcolumntype{s}{>{\hsize=.40\hsize}X}
\newcolumntype{y}{>{\hsize=.25\hsize}X}

\usepackage{inconsolata}

\usepackage{etoolbox}

\AtBeginEnvironment{monoquote}{\fontfamily{cmtt} \selectfont}

\usepackage{float} 
\newfloat{floatquote}{tbp}{loq}
\floatname{floatquote}{Excerpt}
\usepackage{caption}
\usepackage{cleveref}
\crefname{floatquote}{quote}{quotes}
\Crefname{floatquote}{Quote}{Quotes}

\usepackage{subfiles}
\usepackage{array}
\usepackage{threeparttable}
\usepackage{booktabs}
\usepackage{siunitx} 
\usepackage{amssymb}

\usepackage{ifthen}
\usepackage{color}
\definecolor{myorange}{RGB}{240, 96, 0}

\usepackage[sort&compress,numbers]{natbib}
\title{Scalable Qualitative Coding with LLMs: Chain-of-Thought Reasoning Matches Human Performance in Some Hermeneutic Tasks
\thanks{Code and data available at \url{https://osf.io/k4fg9}}
}

\author{
  Zackary Okun Dunivin \\
  Center for Complex Networks and Systems Research\\
  Luddy School of Informatics, Computer Science, and Engineering\\
  Department of Sociology\\
  Indiana University \\
  Bloomington, Indiana, US\\
  \texttt{zdunivin@iu.edu} \\
}

\begin{document}
\maketitle
\newcommand{\abstractopt}{abstract2} 
\ifthenelse{\equal{\abstractopt}{abstract1}}{ 
  \begin{abstract}
Content analysis, or qualitative coding, extracts meaning from text to discern patterns across a corpus of texts. Recent advances in the interpretive abilities of large language models (LLMs) offer potential for automating the coding process (applying category labels to texts), thereby enabling human researchers to concentrate on more creative research aspects while delegating these interpretive tasks to AI. However, it is critical to ensure machine performance aligns with human understanding before fully relying on AI for this task. This paper is a practical guide for adapting a human-developed codebook for an LLM and assessing its performance. Our methodology incorporates the LLM into the standard workflow, effectively treating it as an additional human coder. Beginning with a codebook and gold standard set of code labels, we ``train'' the LLM to identify the codes, refining code descriptions to improve LLM performance. We then validate the LLM's performance against the human-derived gold standard. Employing this method, GPT-4 with zero-shot prompts gives excellent intercoder reliability (Cohen's $\kappa \geq 0.79$) for 3 of 9 codes in our socio-historical case study, and substantial reliability ($\kappa \geq 0.6$) for 8 of 9 codes. In contrast, GPT-3.5 greatly underperforms for all codes ($\mu(\kappa) = 0.34$). Importantly, we find that coding performance improves when the LLM is prompted to provide rationale justifying its coding decisions. Our results indicate that for certain codebooks, current state-of-the-art LLMs are already adept at large-scale content analysis. Furthermore, they suggest the next generation of models will likely render AI coding a viable option for a majority of codebooks.

\end{abstract}
}{
\begin{abstract}
Qualitative coding, or content analysis, extracts meaning from text to discern quantitative patterns across a corpus of texts. Recently, advances in the interpretive abilities of large language models (LLMs) offer potential for automating the coding process (applying category labels to texts), thereby enabling human researchers to concentrate on more creative research aspects, while delegating these interpretive tasks to AI. Our case study comprises a set of socio-historical codes on dense, paragraph-long passages representative of a humanistic study. We show that GPT-4 is capable of human-equivalent interpretations, whereas GPT-3.5 is not. Compared to our human-derived gold standard, GPT-4 delivers excellent intercoder reliability (Cohen's $\kappa \geq 0.79$) for 3 of 9 codes, and substantial reliability ($\kappa \geq 0.6$) for 8 of 9 codes. In contrast, GPT-3.5 greatly underperforms for all codes ($mean(\kappa) = 0.34$; $max(\kappa) = 0.55$). Importantly, we find that coding fidelity improves considerably when the LLM is prompted to give rationale justifying its coding decisions (chain-of-thought reasoning). We present these and other findings along with a set of best practices for adapting traditional codebooks for LLMs. Our results indicate that for certain codebooks, state-of-the-art LLMs are already adept at large-scale content analysis. Furthermore, they suggest the next generation of models will likely render AI coding a viable option for a majority of codebooks.

\end{abstract}
}


\section{Introduction}
Text categorization, commonly referred to as content analysis and qualitative coding in the social sciences, plays an important role in scholarly research and industrial applications. This process traditionally relies on human expertise to interpret the nuanced and often complex meanings embedded in texts \cite{strauss1967discovery,saldana2009coding}. The difficulty lies in the multifaceted nature of meaning and the challenge of fitting real-world complexity into discrete categories, even for skilled readers. Historically, these challenges have positioned text categorization as a task unsuitable for machine learning approaches \cite{malik2020hierarchy}, despite robust attempts \cite{nelson2020computational,dhar2021text}.

Recent developments in artificial intelligence, notably the advent of transformers with billions of parameters known as large language models (LLMs), have begun to challenge this notion. These models demonstrate increasing capabilities in knowledge, interpretation, reasoning, and creativity expressed in natural language, approaching or even surpassing human performance \cite{bubeck2023sparks, romera2023mathematical,team2023gemini}. The processing speed of artificial intelligence opens up the possibility of categorizing vast quantities of text, far exceeding the limitations of human coding teams restricted to smaller samples. Yet, this opportunity raises a critical question: how can we ensure and maintain the accuracy of machine categorization at a level comparable to human standards?

This study provides the strongest evidence to date that machines are capable of human-quality interpretations of text for the purposes of qualitative coding. Additionally, our report serves as a practical guide to employing LLMs in text categorization and as a reference for those encountering machine-assisted qualitative coding in empirical research. We contribute to the growing body of work that builds confidence in the rigor of LLM-based text categorization \cite{xiao2023supporting, chew2023llm,dai2023llm,tai2023examination}, a field that will expand as these models continue to evolve. Our report emphasizes the redesign of codebooks---comprising category descriptions and coding instructions---specifically for LLMs. We demonstrate how the structure of prompts, the specific requests made to the generative model for categorizing passages, significantly impacts coding fidelity. Even as these models continue to rapidly improve, we expect most of the principles of prompt design we report will remain useful and informative as methodologists explore new models and empiricists automate their coding workflows. Our results are presented through narratives detailing our approach and highlighting potential challenges and demonstrated by LLM-generated analyses to a human-derived gold standard. A summary of best practices for content analysis with an LLM is also presented in tabular format for quick reference.

Key findings of our study include:

\begin{itemize}
\item GPT-4 exhibits human-equivalent performance with zero-shot prompts. 8 of 9 tasks exceed the 0.6 threshold for substantial agreement using Cohen's $\kappa$. 3 of 9 tasks exceed the 0.75 threshold for excellent agreement.
\item GPT-3.5, when given the same prompts, has an average intercoder reliability of 0.34 across all codes.
\item Codebooks designed for human coders need reworking for LLM application, requiring iterative manual testing to refine phrasing and improve model comprehension.
\item Agreement improves when the LLM provides rationale for code assignments: $\mu(\kappa) = 0.68$ vs. $\mu(\kappa) = 0.59$.
\item Agreement improves when presenting each code as a separate prompt, rather than the codebook as a whole: $\mu(\kappa) = 0.68$ vs. $\mu(\kappa) = 0.60$.
\end{itemize}

\subsection{Automating Content Analysis: Past and Present}
Prior work on automating content analysis entailed training machine learning models on large quantities of text. Supervised models, typically some form of linear regression, learn to associate text features with user-specified categories \cite{kadhim2019survey}. This process captures half the traditional human-coded process by using human-derived codes and examples, but fails to leverage abstract code descriptions found in a codebook, as well as requiring large quantities of human-annotated data. Unsupervised models, such as LDA \cite{blei2003latent,jelodar2019latent} or BERTTopic \cite{grootendorst2022bertopic}, develop their own categorizations from unlabeled training sets. This process does not require time-intensive labeling, but rarely captures the specific categories that a researcher intends to target.

The latest generation of LLMs (e.g, GPT \cite{brown2020language}, LLaMa \cite{touvron2023llama}, Mistral \cite{mistral2023mixtral},
Claude \cite{anthropic2023claude}) differ notably from previous machine learning models in that they can perform new tasks specified through natural language prompts. A user can specify a task that the model was not trained on, give few (single digit) or no examples, and the model will return output conforming to the specifications. Demonstrated successes include computer code generation \cite{nowakowski2024ai}, creative writing \cite{gomez2023confederacy}, and quantitative reasoning \cite{bischoff2024ai}. We are only beginning to understand and expand upon the limitations of these models. By converting natural language requests into highly intelligent output across vast and indeterminate domains, LLMs lower the technical barriers to machine learning by making its application more naturalistic and eliminating the need for large training data. Beyond this, LLM's capacities in many domains far exceed the specialized machine learning models that preceded them, suggesting that for many applications, including scholarly inquiry, artificial intelligence is overwhelmingly more accessible and capable in 2024 than it was just two years prior.

Early studies of content analysis with LLMs are encouraging. Xiao et al. \cite{xiao2023supporting} demonstrate moderate success, Cohen's $\kappa = 0.61$ and $\kappa = 0.38$, in two linguistic tasks using GPT-3. Chew et al. \cite{chew2023llm} report high success on many of 19 tasks across three datasets, and results that are indistinguishable from random for others. It is difficult to evaluate their results due to the choice of Gwet's AC1, which is biased toward agreement on negative codings rather than positive, whereas most standard measures of intercoder reliability do the opposite \cite{vach2023gwet}. However, we commend Chew et al.'s approach to adapting codebooks for LLMs, which is communicated with great detail and clarity. A survey of 20 empirical pieces reports ``mixed-results'' of using GPT-3 to automate ``text annotation,'' a term that ties their framing to ``data annotation'', labeling data for in machine learning \cite{ollion2023chatgpt}, rather than content analysis in the tradition of grounded theory \cite{strauss1997grounded}.

We present here three advances to these studies. 1) We report the first methodological account of automating qualitative coding using GPT-4, which, along with other recent models, greatly improves upon many of GPT-3's capabilities \cite{openai2023gpt,yuan2024selfrewarding,mistral2023mixtral}. 2) We provide the first conclusive evidence that LLMs are capable of human-equivalent performance in qualitative coding, and do so on larger passages of text, wherein meaning is often woven through multiple interrelated clauses. 3) We demonstrate that GPT is better at interpreting text when it is tasked with justifying its coding decisions (chain-of-thought prompting) rather than applying codes without an accompanying explanation.


\subsection{Case Study: W.E.B. Du Bois's Characterization in News Media}
In order to present a realistic challenge of using an LLM to do qualitative coding, we make a case study of our own work. We adapted a codebook written by the authors to understand how the scholar and activist W.E.B. Du Bois has been characterized in news media over time. The codebook is composed of 9 codes in 3 categories. Due to multiple layers of agency (who is doing what) and voice (who is saying what), the tasks are difficult even for human interpreters. Applying the codes is also complicated because it can be difficult to differentiate Du Bois's scholarship from his political activism, as Du Bois's theoretical contributions have profound implications for understanding race and the social-historical position of Black persons in the United States and beyond, making them powerful activist tools. We are particularly interested in understanding how different facets of Du Bois's activities contributed to his canonization in the public imagination as the preeminent figure for understanding Black political struggle. Table \ref{table:code_listing} gives the codes in brief. Complete examples of the original human and modified-for-GPT codebook are included in the appendix.

The training and test data for our study were random samples of passages from New York Times articles (1970--2023) that mention W.E.B. Du Bois. 232 passages were automatically extracted as concurrent paragraphs containing ``Du Bois''. The average number of words was 94 ($\sigma=70$), and the average number of sentences was 3.75 ($\sigma=2.88$). To give a better sense of the size of our passages, this paragraph has 76 words across 4 sentences.

\subfile{tables/abbr_codebook.tex}

\section{Results}
\label{sec:results}
\subsection{Adapting a Codebook for an LLM}
Initially, we developed a codebook for human coders using standard methods. This process involved exploratory reading to define and refine codes. Codes were derived to probe particular substantive hypotheses, some of which preceded our exploratory reading, and others which resulted from it. We then applied these codes iteratively, adjusting them as needed until we achieved high intercoder reliability with a test set. We adapted these code descriptions for use with a large language model, evaluating the LLM's performance on a training set of text passages. Where we found ambiguities or deficiencies in the model's interpretation, we refined the code descriptions accordingly. This iterative process of definition, evaluation, and refinement follows Nelson's Computational Grounded Theory paradigm \cite{nelson2020computational}, the core of which is common to all qualitative code development processes whether or not coding is automated \cite{strauss1997grounded}.

Our experience modifying the code descriptions yielded several key insights related in the following paragraphs. We encourage readers interested in a fuller account of this process to read Chew et al.'s study \cite{chew2023llm} describing their process of LLM-Assisted Content Analysis (LACA), which relates a process similar to our own in greater detail.

\paragraph{LLM-generated rationale are essential for evaluating performance.}
In adapting the codebook, we wanted to understand not just which codes the model struggled to interpret correctly, but what aspects of the code the model failed to capture. To achieve this, we structured our prompts to require GPT to justify its decision to apply or not apply each code. These rationale were invaluable. They often highlighted parts of the code description that were ambiguous or imprecisely defined, leading the model to misinterpret them. Whenever a rationale repeatedly pointed to such an issue, we revised the corresponding code. We then retested the passage to check that the code was correctly applied and the rationale aligned with our intended interpretation of the code. Sometimes a revision would not improve the interpretations for the passages in question; other times it would fix the interpretations for those passages, but would introduce new problems in passages which were previously coded correctly.

\begin{figure}
    \centering
    \includegraphics[width=\textwidth]{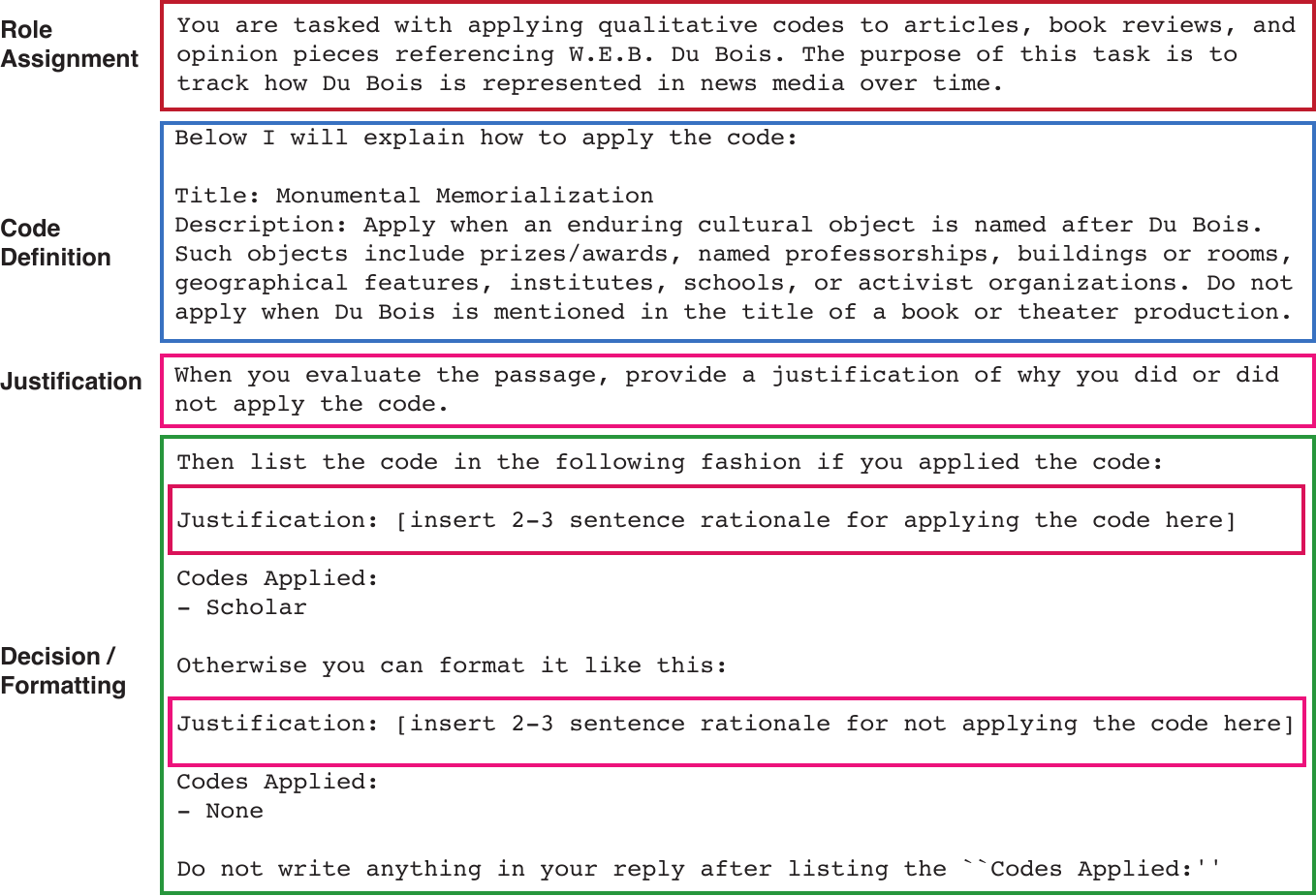}
    \caption{The chain-of-thought prompt sequence.}
    \label{fig:prompt-example}
\end{figure}

Figure \ref{fig:prompt-example} demonstrates an effective method of prompting GPT to provide rationale for its code selections. The initial instruction is given by the Justification section of the prompt, and solicited again in the Decision/Formatting box.

\begin{figure}
    \centering
    \includegraphics[width=\textwidth]{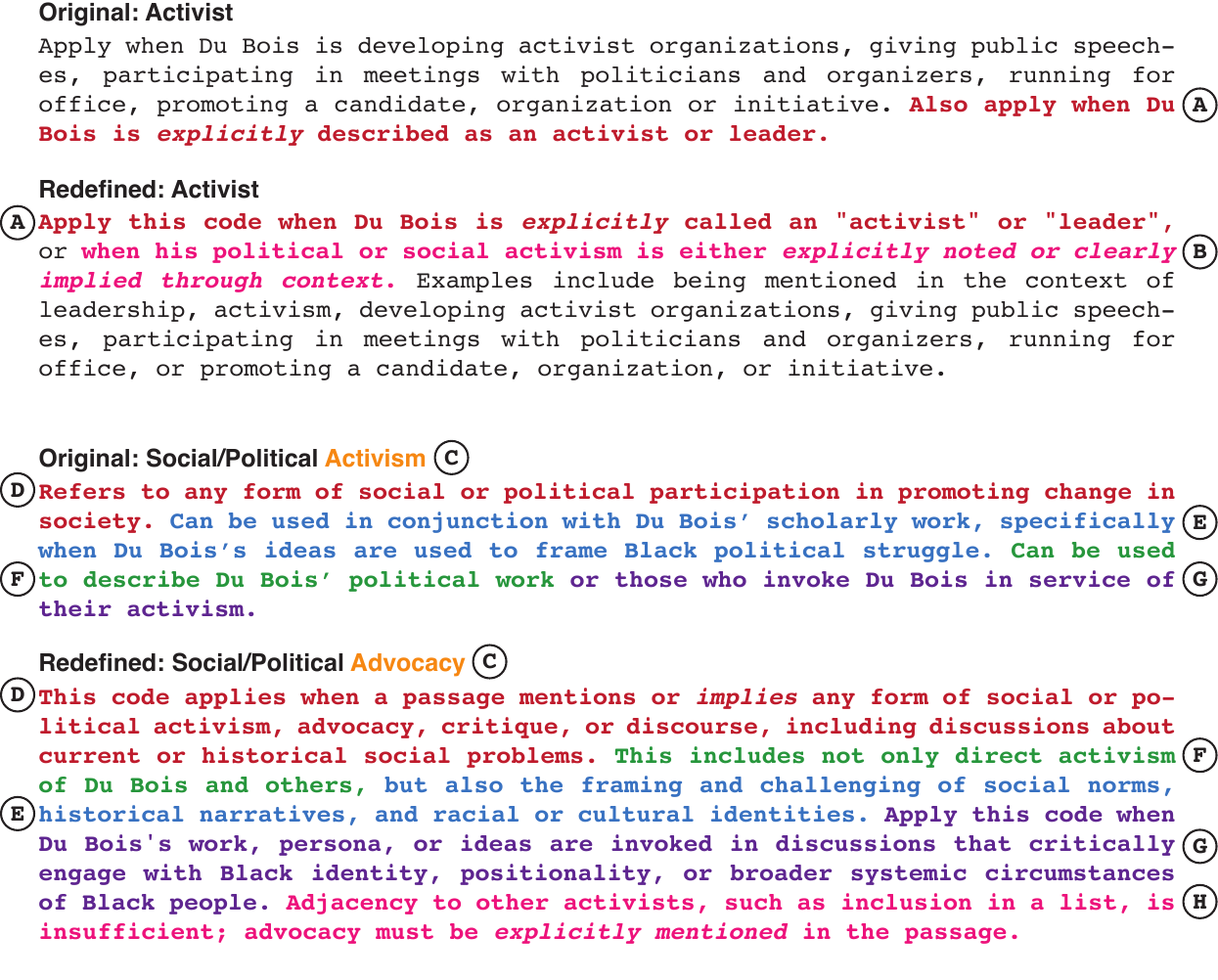}
    \caption{Two examples of prompt redefinition. Colored, alphabetically labeled blocks of text show alterations derived through iterative code refinement. Italics draw attention to direction to constrain interpretive scope to implicit or explicit information.}
    \label{fig:prompt_edits}
\end{figure}

\paragraph{LLMs require more precise descriptions than do human readers.}
Human coders do not rely solely on a written codebook. Their interpretation of the codes is enriched through the codebook development process, discussions with fellow coders, and supplementary oral instructions. An LLM lacks this interactive and historical context and must interpret codes entirely from written descriptions. Our work modifying the codebook for GPT revealed information that, while implicitly understood by the code developers, wasn't explicitly stated in the code descriptions. This process not only aided in refining the codebook for automated coding, but also improved our own understandings of the codes. This ultimately led to clearer definition of the codes, thereby enhancing future manual coding processes as well. Figure \ref{fig:prompt_edits} demonstrates how the Monumental Memorialization and Social/Political Advocacy codes were redefined to improve GPT's comprehension.

Often, we encountered cases where ambiguous phrasing was obvious to humans, but challenging for the LLM. Our codebook contains two codes that relate to Du Bois's reputation among academics and activists. These codes are meant to evaluate whether Du Bois appears in a news story because either an academic or activist mentioned him. Initially, we titled this code ``Academic Repute,'' which worked well for human coders. GPT, however, consistently misinterpretted this code as pertaining to Du Bois's esteem \textit{as} an academic, rather than \textit{among} or \textit{by} academics (the meaning of ``among'' remains ambiguous even here). We tried numerous iterations this code without success. Nevertheless, altering the title of the code to the far more literal ``Out of the Mouth of Academics'' dramatically improved performance, even when paired with the original code description. In another case, the code titled ``Social/Political Activism'' was revised to ``Social/Political Advocacy'' (Figure \ref{fig:prompt_edits}, C) because GPT did not consider social critique to be a form of activism, even when it was specifically instructed to.

We found that words indicating how much the model should draw on context or its own outside knowledge had large impacts on the model's outputs, often to the desired effect. In particular, instructing the model to restrict itself to ``explicit'' meanings, or to draw on ``implicit'' meanings, often helped the model with part of a code description it had struggled with. Figure \ref{fig:prompt_edits} B and D demonstrate the addition of such verbiage to control scope. 

Both mandatory (do) and prohibitory (do not) phrasing were observed by the model, though mandatory phrasing seemed more successful, a finding reported by other researchers \cite{bsharat2023principled}. The ordering of directives also impacted how likely the model was to follow them. We found that moving a phrase that was ignored in the coding rationale toward the front of the definition made the model more likely to follow its specifications, as in Figure \ref{fig:prompt_edits} A. When a very specific problem was observed repeatedly, it was sometimes necessary to add a directive to correct it, as in Figure \ref{fig:prompt_edits} H.

\paragraph{Prompting for machine-readable output.} To fully automate the coding process, model output must be reliably readable by a computer. The LLM generates text, which must be interpreted by another script into a data structure, such as a table, for further analysis. Instructing exactly how to format the output produced machine-readable results with GPT-4 and GPT-3.5. Critically, this involves specifying a tag that the interpreting script locates, after which follows a reliably formatted list of codes. The Decision/Formatting component Figure \ref{fig:prompt-example} illustrates how to constrain model output and produce consistent results across queries. Additionally, because GPT tends to be excessively verbose and summarize its output, particularly at higher temperatures, we informed the model that we do not want any output to follow the code list.

\subsection{Selecting a Model and Writing Prompts for Optimal Performance}
Once the code descriptions have been revised for LLM text categorization, numerous other decisions remain about how to prompt a model to execute the content analysis. We present these as a separate step for the sake of clarity, but in reality, we developed our approach in iteratively and tandem with revising the code descriptions. We hope future methodologists and empiricists will benefit from what we learned during this process, and that less exploration of these components will be necessary so practitioners can focus on application or exploration of calibrations not explored here. We summarize all our recommendations for qualitative coding with an LLM in Table \ref{table:principles}.

There is large and growing body of academic and nonacademic literature on prompt engineering: constructing user-defined input to elicit the best model output. In fact, the codebook adaptation in the previous section was in large part an exercise in prompt engineering. However, in this second section, prompt engineering refers more to the broader context of task description than the code definitions. In this section we report how different prompts influence the quality of machine categorization. Additionally we compare performance when the LLM is tasked with assigning each code independently to when the model is given the full codebook and assigned with coding all 9 codes as a single task. We refer to these as the ``Per Code'' and ``Full Codebook'' approach respectively.

Studies have shown that LLM decision-making improves when the model is prompted to account for its decisions \cite{wei2022chain,madaan2022text}. This is generally known as chain-of-thought (CoT) prompting or reasoning, and refers to breaking down tasks into specific components, one or more of which involve planning for future steps or reflection on previous ones. Our prompts, which can be viewed in full in the appendix, apply chain-of-thought prompting by including 1) a role assignment step, informing the machine of its purpose, 2) a task description step, specifying the code definition, 3) a justification step, instructing the model to provide a rationale for its decision, and 4) a decision step, wherein the model delivers its ultimate analysis in a consistent, machine-readable format. An example of the chain-of-thought prompt sequence is given by Figure \ref{fig:prompt-example}.

We use zero-shot prompts throughout this study. Zero-shot refers to providing the model only the task description, without giving examples of correctly executed responses. Xiao et al. found few-shot prompting improves coding and performance on other tasks \cite{xiao2023supporting}, whereas Chew et al. largely employed zero-shot prompts \cite{chew2023llm}. Our case study involves evaluating paragraph-long passages rather than single clauses. We found that information in the examples was drawn upon by the LLM and interfered with its coding decisions. We also found that in Full Codebook prompts, giving examples greatly expanded the prompt, negatively impacting results. We suggest that when content is more literary or historical, zero-shot prompts are probably preferred, but that most coding tasks will benefit from few-shot prompting as demonstrated by the results of many other studies across domains.

Performance comparisons are relative to the human-derived gold standard hidden from the LLM at all stages of development. We used the default settings for the GPT API where temperature is set to 0 and nucleus sampling (\texttt{top\_p}) is set to 1. We specified the task description as a ``system prompt'', and provide each passage as a ``user prompt''. A system prompt gives the LLM its purpose, clearly specifying the task it is meant to address, whereas a user prompt provides the input to which the model responds by generating output. We did not investigate whether intercoder agreement suffers with the default system prompt, while combining the task description and passage as a user prompt.

\subfile{tables/fullest_comparison.tex}

\paragraph{GPT-4 greatly outperforms GPT-3.5.}
We found that GPT-4 approaches human performance for 3 codes: Activist: $\kappa$ = 0.81; Monumental Memorialization: $\kappa$ = 1.00; Collective Synecdoche $\kappa$ = 0.79. GPT-4 prompted for rationale provides considerably higher quality code assignments than GPT-3.5, except in the case of the Out of the Mouth of Activists code, which no configuration handled well. It is especially notable that GPT-4 and GPT-3.5 differed in their most accurately interpreted codes. In the 3 tasks GPT-4 executed best, GPT-3.5's performance was slightly below its own average, $mean(\kappa_{all})$ = 0.34 vs. $mean(\kappa)$ = 0.32.

\paragraph{Coding fidelty improves when codes are presented as individual tasks.}
We adapted our codebook by presenting the entire codebook to GPT along with task instructions. However, we found in testing that performance improved when GPT was given each task independently. This ``per code'' approach was taken by one recent study exploring content analysis with non-mutually exclusive codes (permitting multiply coded passages) \cite{chew2023llm}, but not another, which tested only two codes  \cite{xiao2023supporting}. Table \ref{table:fullest_comparison} compares the GPT-4 performance when presented individual tasks for each code (``Per Code'') and when presented all tasks in a single prompt (``Full Codebook''). We found that for the 3 human-equivalent tasks (Activist, Monumental Memorialization, and Collective Synecdoche) the Per Code performance far exceeded the Full Codebook for 2 tasks, and was comparable for 1. For 2 other tasks, Mention of Scholarly Work and Coalition Building, we found that the Per Code configuration produced considerably higher agreement, whereas Full Codebook performed comparably to the Per Code in the remaining 4 tasks.

\paragraph{Coding fidelity improves when the model is prompted to justify its coding decisions.} Consistent with other experiments with chain-of-thought (CoT) reasoning in LLMs, we found that coding agreement benefited strongly from prompting the model to explain itself \cite{chu2023survey}.
Table \ref{table:fullest_comparison} shows the effect of prompting for rationale on three pairs of conditions: Per Code GPT-4, Full Codebook GPT-4, and Per Code GPT-3.5. We found that across all codes and conditions, with one exception, CoT prompting produces higher or equivalent intercoder reliability with the gold standard. Using GPT-4, average Per Code agreement improved from 0.59 to 0.68, and average Full Codebook agreement improved from 0.46 to 0.60. Furthermore, a majority of pairs showed substantial improvement when the codes were assigned after providing reasoning for coding decisions.

\subfile{tables/principles.tex}

\section{Discussion}

\paragraph{Determining appropriate domains for LLM-assisted qualitative coding.} Previous methods of automated text categorization, both supervised and unsupervised, rarely met the standards of traditional social scientists and humanists, and were instead generally employed by data scientists. Capturing meaning, particularly complex meaning, through machine learning has largely been an elusive goal \cite{malik2020hierarchy}. Despite our own former skepticism, we predict that LLMs will be capable of applying most qualitative codebooks within the year. However, our results show that even within the scope of a single codebook, interpretation quality varies. Thus, different disciplines and domains should expect model success and the ease of transitioning a codebook to vary considerably. We suspect that more humanistic and ``softer'' scientific approaches will (continue to) be more resistant to machine interpretation than problems posed by scholars who identify with ``harder'' sciences, to say nothing of their ability to convince their peers of its validity. We do not oppose developing evaluation benchmarks for qualitative coding to assess which models are adept at what variety of task, but neither do we advocate it; meaning is manifold and emergent, and much of its beauty derives from its resistance to reduction and definition. Instead, we suggest those who wish to employ an LLM to perform content analysis survey similar attempts and simply experiment on their own. The process of discovering triumphs, workarounds, and limitations of working with these models was not only fascinating, but tremendously fun.

\paragraph{Practical aspects of transitioning to content analysis with LLMs.}
While artificial intelligence potentially opens up much larger datasets to qualitative scholars, there is still a considerable technical barrier to automating content analysis. Development of an LLM adapted codebook is feasible for anyone regardless of technical skill by interacting with an LLM through chat-like Web platforms provided by proprietary model developers. 
However, systematically testing prompts or applying a completed codebook to the full dataset requires moderate skill in writing scripts in a language such as Python. Rather than suggest that all scholars become programmers, we encourage researchers to develop partnerships with students or community members seeking programming or research experience as a form of project-based education. Conversely, we suggest that data scientists pursue partnerships with traditional social scientists and humanists, who are often better positioned to develop coding schema to flush out complex meanings embedded in text, which are now more tractable to machine learning.

\paragraph{Handling passages where model interpretation is poor.}
Overwhelmingly, GPT-4's interpretations were accurate and human-like. However, we found repeatedly that GPT-4, like a human reader, struggled with edge cases, especially where implicit information was required to make a judgment. We are encouraged by this finding, and argue that with automated analysis, fidelity is less important than it is with humans. Because statistical power increases with the number of observations, noise is more tolerable in machine-applied codes, as automated coding potentially increases sample size by orders of magnitude. Notably, this assumes that error is restricted to edge cases and is not otherwise systematically biased. We also advise against automated coding where datasets are small, as in interviews, where it is likely as efficient to code entirely by hand. As models improve and can provide confidence estimates for their statements \cite{lin2023generating,zhou2023navigating,chen2024introducing}, ML content analysis workflows should include manual review of passages with uncertain code assignments. Anecdotally, we found that GPT-4 could intelligently reflect on its responses when prompted to do so. When presented the output of another model instance, GPT-4, acting as an untrained ``critic'' model \cite{paul2023refiner}, was often able to identify when it had encountered an edge case without prompting, as well as recognize and revise obvious mistakes. Our experiences suggest that a human-in-the-loop tag-for-manual-review workflow or a two-step automated reflect-and-revise workflow may already be feasible with GPT-4 and similar models.

\section{Conclusion}
Our results using state-of-the-art models lead us to recommend that scholars who do much qualitative coding consider automated coding with LLMs a potentially viable option today. We especially encourage skeptics to probe these tools' capacities, as it is useful to know their limitations. Over the next year, models such as those incorporating memory \cite{shinn2023reflexion}, (multi-)agential models that dialogue and revise prior to rendering output \cite{yao2022react, xu2023reasoning}, and architectures that can handle larger inputs \cite{gu2023mamba}, will almost certainly greatly improve upon GPT-4's current abilities. When those models are made available, researchers who have already experimented with LLMs will be best positioned to make use of the new tools. The efficiency of automation is compelling, but we are most enthusiastic about the ability to probe much larger datasets than ever before, potentially illuminating patterns too rare or too fuzzy to detect with a sample numbering in the tens or hundreds rather than thousands or beyond.

\section*{Acknowledgments}
The author thanks Harry Yan, Pat Wall, Patrick Kaminski, Adam Fisch, Alicia Chen, and Francisco Muñoz for their helpful comments toward improving this manuscript. I am especially grateful to Tania Ravaei for collaborating on codebook development.


\bibliography{references}

\end{document}

%% file: tables/abbr_codebook.tex
\def\arraystretch{1.4}

\begin{table}[ht]
\centering
\caption{Categories and descriptions for 9 codes.} 
\begin{tabularx}{\linewidth}{sk}
\toprule
\multicolumn{2}{l}{\bf Characterization of Du Bois}\\
\cmidrule(r){1-1}
\em Scholar & Describes Du Bois as a scholar or intellectual.\\
\em Activist & Refers to Du Bois's political or social activism.\\

\multicolumn{2}{l}{\bf General Themes}\\
\cmidrule(r){1-1}
\em  Monumental Memorialization & Refers to an enduring cultural object named after Du Bois.\\
\em Mention of Scholarly Work & Mentions or quotes specific academic works by Du Bois.\\ 
\em  Social/Political Advocacy &  Mentions or implies social or political activism, advocacy, or critique.\\

\multicolumn{2}{l}{\bf Canonization Processes}\\
\cmidrule(r){1-1}
\em Coalition Building & Refers to Du Bois’s activities with activist or academic organizations.\\
\em Out of the Mouth of Academics & Describes an academic organization engaging with Du Bois’s legacy.\\
\em Out of the Mouth of Activists & Describes an activist organization engaging with Du Bois’s legacy.\\
\em Collective Synecdoche & Mentions Du Bois alongside other figures in order to represent some facet of a culture, era, or ideology.\\
\bottomrule
\label{table:code_listing}
\end{tabularx}
\end{table}

%% file: tables/fullest_comparison.tex
\bgroup
\def\arraystretch{1.4}
\begin{table}[!htbp]
\centering
\caption{Intercoder reliability (Cohen's $\kappa$) for all codes on 111 gold standard passages. Best overall performance is shown in bold.  Italics indicate the highest intercoder reliability between pairs with and without prompting for rationale (CoT vs. No CoT); if the pair are equivalent neither is italicized. Two values are considered equivalent if their difference does not exceed 0.02.} 
\centering 

\newcommand*{\theader}[1]{\multicolumn{1}{c}}
\newcommand\myrepeat[2]{%
  \begingroup
  \lccode`m=`#2\relax
  \lowercase\expandafter{\romannumeral#1000}%
  \endgroup
}

\newcommand{\subit}[1]{\textit{\textsubscript{#1}}}

\newcommand{\sig[1]}{\myrepeat{#1}{*}}

\newcommand{\stdlineadjust}{\vspace{-1ex}}
\newcommand{\stdline}[2]{& (#1) & (#2)\\}

\sisetup{
    detect-all,
    round-integer-to-decimal = true,
    group-digits             = true,
    group-minimum-digits     = 3,
    group-separator          = {\,},
    table-align-text-pre     = false,
    table-align-text-post    = false,
    input-signs              = + -,
    input-open-uncertainty   = ,
    input-close-uncertainty  = ,
    retain-explicit-plus
}

\begin{threeparttable}
\begin{tabular}{l S[table-format=2]  *6{S[table-format=2.2, 
    table-space-text-post = {$^{**}$}, 
    round-mode=places,
    round-precision=2]}}
\toprule 

\textit{} & \multicolumn{1}{c}{} & \multicolumn{4}{c}{\bf GPT-4} & \multicolumn{2}{c}{\bf GPT-3.5}\\
\cmidrule(r){3-6} \cmidrule(r){7-8}

\textit{} & & \multicolumn{2}{c}{\bf Per Code} & \multicolumn{2}{c}{\bf Full Codebook} & \multicolumn{2}{c}{\bf Per Code}\\
\cmidrule(r){3-4} \cmidrule(r){5-6}  \cmidrule(r){7-8}
\multicolumn{1}{c}{\bf Code} & \multicolumn{1}{c}{\bf Count} & \multicolumn{1}{c}{\bf CoT} & \multicolumn{1}{c}{\bf No CoT} & \multicolumn{1}{c}{\bf CoT} & \multicolumn{1}{c}{\bf No CoT} & \multicolumn{1}{c}{\bf CoT} & \multicolumn{1}{c}{\bf No CoT} \\
\cmidrule(r){1-1} 
\cmidrule(r){2-2} \cmidrule(r){3-3}
\cmidrule(r){4-4}
\cmidrule(r){5-5}
\cmidrule(r){6-6}
\cmidrule(r){7-7}
\cmidrule(r){8-8}
Scholar & 27 & \bf 0.608587 & 0.516202 & \bf 0.592511 & 0.420582 & \em 0.293830 & 0.206271 \\
Activist & 23 & \bf 0.811087 & 0.649161 & \em 0.670021 & 0.616107 & \em 0.390687 & 0.315357 \\
Monumental Memorialization & 13 & \bf 1.000000 & 0.912873 & \em 0.752784 & 0.477237 & 0.287319 & 0.308331 \\
Mention of Scholarly Work & 24 & \bf 0.711960 & \bf 0.690377 & \em 0.521036 & 0.439394 & 0.332384 & \em 0.389061 \\
Social/Political Advocacy & 51 & \bf 0.636185 & 0.603314 & 0.598619 & 0.595025 & \em 0.548560 & 0.505253 \\
Coalition Building & 9 & \bf 0.597201 & 0.443609 & \em 0.433260 & 0.131455 & \em 0.330519 & 0.169576 \\
Out of the Mouth of Academics & 30 & \bf 0.626734 & \bf 0.646272 & \bf 0.649289 & 0.618557 & \em 0.374207 & 0.327899 \\
Out of the Mouth of Activists & 11 & \em 0.304274 & 0.091818 & \bf 0.341246 & 0.177778 & \em 0.209709 & 0.085274 \\
Collective Synecdoche & 26 & \bf 0.788123 & 0.776959 & \bf 0.809278 & 0.706479 & 0.274712 & 0.274712 \\
\midrule
Mean & 24 &  \bf 0.676017 & 0.592287 & \em 0.596449 & 0.464735 & \em 0.337992 & 0.291278 \\

\bottomrule 
\end{tabular}
\end{threeparttable}
\label{table:fullest_comparison} 
\end{table}
\egroup

%% file: tables/principles.tex
\def\arraystretch{1.4}

\begin{table}[ht]
\centering
\caption{Principles of prompting an LLM for qualitative coding.} 
\begin{tabularx}{\linewidth}{yk}
\toprule
\multicolumn{2}{l}{\bf Task Instructions}\\
\cmidrule(r){1-1}
\em Prompt for Rationale & Model fidelity improves when instructed to justify its coding decisions.\\
\em One Task Per Code & Model fidelity improves when given each code as a separate task.\\
\em Brevity & Shorter task descriptions are more likely to be faithfully executed by the model.\\
\em Structured Output & Instruct the model to format its output to ensure uniform responses.\\
\multicolumn{2}{l}{\bf Code Definitions}\\
\cmidrule(r){1-1}
\em Word Choice & A single high-content word can be changed to align with the LLM's built-in ontology.\\
\em Clause Order & Clauses are more likely to be observed when introduced earlier in the code description.\\
\em Mandates/Prohibitions & Both can be effective, but it is easier to get the model to ``do'' than ``do not''.\\
\em Code Titles & Altering the code title can have a large effect even without altering the definition.\\
\em Interpretation Scope & Use words like ``implicit'' and ``explicit'' when interpretation is too limited or expansive.\\

\multicolumn{2}{l}{\bf Chain-of-Thought Prompt Sequence}\\
\cmidrule(r){1-1}
\em 1. Role Assignment & Supply the model its purpose, e.g., "You will be applying category labels to passages."\\
\em 2. Code Definition & Provide the code title(s) and description(s).\\
\em 3. Justification & Request that the model provide evidence of its reasoning.\\
\em 4. Decision & Instruct the model to list the codes that apply to the passage in a consistent format.\\
\bottomrule
\label{table:principles}
\end{tabularx}
\end{table}

%% file: main.bbl
\begin{thebibliography}{10}

\bibitem{strauss1967discovery}
Anselm~L Strauss.
\newblock 1967.
\newblock {\em The discovery of grounded theory: Strategies for qualitative research}.
\newblock Aldine.

\bibitem{saldana2009coding}
Johnny Salda{\~n}a.
\newblock 2009.
\newblock {\em The coding manual for qualitative researchers}.
\newblock SAGE Publications.

\bibitem{malik2020hierarchy}
Momin~M Malik.
\newblock 2020.
\newblock A hierarchy of limitations in machine learning.
\newblock {\em arXiv preprint arXiv:2002.05193}.

\bibitem{nelson2020computational}
Laura~K Nelson.
\newblock 2017.
\newblock {Computational Grounded Theory}: {A} methodological framework.
\newblock {\em Sociological Methods \& Research}, 49(1):3--42.

\bibitem{dhar2021text}
Ankita Dhar, Himadri Mukherjee, Niladri~Sekhar Dash, and Kaushik Roy.
\newblock 2021.
\newblock Text categorization: past and present.
\newblock {\em Artificial Intelligence Review}, 54:3007--3054.

\bibitem{bubeck2023sparks}
S{\'e}bastien Bubeck, Varun Chandrasekaran, Ronen Eldan, Johannes Gehrke, Eric Horvitz, Ece Kamar, Peter Lee, Yin~Tat Lee, Yuanzhi Li, Scott Lundberg, et~al.
\newblock 2023.
\newblock Sparks of artificial general intelligence: {E}arly experiments with {GPT}-4.
\newblock {\em arXiv preprint arXiv:2303.12712}.

\bibitem{romera2023mathematical}
Bernardino Romera-Paredes, Mohammadamin Barekatain, Alexander Novikov, Matej Balog, M~Pawan Kumar, Emilien Dupont, Francisco~JR Ruiz, Jordan~S Ellenberg, Pengming Wang, Omar Fawzi, et~al.
\newblock 2023.
\newblock Mathematical discoveries from program search with large language models.
\newblock {\em Nature}, 625:1--3.

\bibitem{team2023gemini}
Gemini Team, Rohan Anil, Sebastian Borgeaud, Yonghui Wu, Jean-Baptiste Alayrac, Jiahui Yu, Radu Soricut, Johan Schalkwyk, Andrew~M Dai, Anja Hauth, et~al.
\newblock 2023.
\newblock Gemini: {A} family of highly capable multimodal models.
\newblock {\em arXiv preprint arXiv:2312.11805}.

\bibitem{xiao2023supporting}
Ziang Xiao, Xingdi Yuan, Q~Vera Liao, Rania Abdelghani, and Pierre-Yves Oudeyer.
\newblock 2023.
\newblock Supporting qualitative analysis with large language models: Combining codebook with {GPT}-3 for deductive coding.
\newblock In {\em Companion Proceedings of the 28th International Conference on Intelligent User Interfaces}.

\bibitem{chew2023llm}
Robert Chew, John Bollenbacher, Michael Wenger, Jessica Speer, and Annice Kim.
\newblock 2023.
\newblock {LLM}-assisted content analysis: {U}sing large language models to support deductive coding.
\newblock {\em arXiv preprint arXiv:2306.14924}.

\bibitem{dai2023llm}
Shih-Chieh Dai, Aiping Xiong, and Lun-Wei Ku.
\newblock 2023.
\newblock {LLM}-in-the-loop: {L}everaging large language model for thematic analysis.
\newblock {\em arXiv preprint arXiv:2310.15100}.

\bibitem{tai2023examination}
Robert~H Tai, Lillian~R Bentley, Xin Xia, Jason~M Sitt, Sarah~C Fankhauser, Ana~M Chicas-Mosier, and Barnas~G Monteith.
\newblock 2023.
\newblock An examination of the use of large language models to aid analysis of textual data.
\newblock {\em bioRxiv preprint bioRxiv:2023.07.17.549361}.

\bibitem{kadhim2019survey}
Ammar~Ismael Kadhim.
\newblock 2019.
\newblock Survey on supervised machine learning techniques for automatic text classification.
\newblock {\em Artificial Intelligence Review}, 52(1):273--292.

\bibitem{blei2003latent}
David~M Blei, Andrew~Y Ng, and Michael~I Jordan.
\newblock 2003.
\newblock Latent {D}irichlet allocation.
\newblock {\em Journal of Machine Learning Research}, 3(1):993--1022.

\bibitem{jelodar2019latent}
Hamed Jelodar, Yongli Wang, Chi Yuan, Xia Feng, Xiahui Jiang, Yanchao Li, and Liang Zhao.
\newblock 2019.
\newblock Latent {Dirichlet} allocation ({LDA}) and topic modeling: {M}odels, applications, a survey.
\newblock {\em Multimedia Tools and Applications}, 78:15169--15211.

\bibitem{grootendorst2022bertopic}
Maarten Grootendorst.
\newblock 2022.
\newblock {BERT}opic: {N}eural topic modeling with a class-based {TF-IDF} procedure.
\newblock {\em arXiv preprint arXiv:2203.05794}.

\bibitem{brown2020language}
Tom Brown, Benjamin Mann, Nick Ryder, Melanie Subbiah, Jared~D Kaplan, Prafulla Dhariwal, Arvind Neelakantan, Pranav Shyam, Girish Sastry, Amanda Askell, et~al.
\newblock 2020.
\newblock Language models are few-shot learners.
\newblock {\em Advances in Neural Information Processing Systems}, 33:1877--1901.

\bibitem{touvron2023llama}
Hugo Touvron, Thibaut Lavril, Gautier Izacard, Xavier Martinet, Marie-Anne Lachaux, Timoth{\'e}e Lacroix, Baptiste Rozi{\`e}re, Naman Goyal, Eric Hambro, Faisal Azhar, et~al.
\newblock 2023.
\newblock {LLaMa}: {O}pen and efficient foundation language models.
\newblock {\em arXiv preprint arXiv:2302.13971}.

\bibitem{mistral2023mixtral}
{MistralAI}.
\newblock 2023.
\newblock Mixtral of experts: {A} high quality sparse mixture-of-experts.
\newblock \url{https://mistral.ai/news/mixtral-of-experts}.
\newblock Accessed: 2024-01-13.

\bibitem{anthropic2023claude}
Anthropic.
\newblock 2023.
\newblock Claude 2.
\newblock \url{https://www.anthropic.com/index/claude-2}.
\newblock Accessed: 2024-01-18.

\bibitem{nowakowski2024ai}
Jan Nowakowski and Jan Keller.
\newblock 2024.
\newblock {AI}-powered patching: {T}he future of automated vulnerability fixes.
\newblock {\em Google Security Engineering Technical Report}.

\bibitem{gomez2023confederacy}
Carlos G{\'o}mez-Rodr{\'\i}guez and Paul Williams.
\newblock 2023.
\newblock A confederacy of models: {A} comprehensive evaluation of {LLM}s on creative writing.
\newblock {\em arXiv preprint arXiv:2310.08433}.

\bibitem{bischoff2024ai}
Manon Bischoff.
\newblock 2024.
\newblock {AI} matches the abilities of the best {Math Olympians}.
\newblock {\em Scientific American}.

\bibitem{vach2023gwet}
Werner Vach and Oke Gerke.
\newblock 2023.
\newblock {Gwet's AC1} is not a substitute for {Cohen's kappa -- A} comparison of basic properties.
\newblock {\em MethodsX}, 10:102212.

\bibitem{ollion2023chatgpt}
Etienne Ollion, Rubing Shen, Ana Macanovic, and Arnault Chatelain.
\newblock 2023.
\newblock Chat{GPT} for text annotation? {M}ind the hype!
\newblock {\em SocArXiv preprint doi:10.31235/osf.io/x58kn}.

\bibitem{strauss1997grounded}
Anselm Strauss and Juliet~M Corbin.
\newblock 1997.
\newblock {\em Grounded theory in practice}.
\newblock SAGE Publications.

\bibitem{openai2023gpt}
{OpenAI}.
\newblock 2023.
\newblock {GPT}-4.
\newblock \url{https://openai.com/research/gpt-4}.
\newblock Accessed: 2024-01-18.

\bibitem{yuan2024selfrewarding}
Weizhe Yuan, Richard~Yuanzhe Pang, Kyunghyun Cho, Sainbayar Sukhbaatar, Jing Xu, and Jason Weston.
\newblock 2024.
\newblock Self-rewarding language models.
\newblock {\em arXiv preprint arXiv:2401.10020}.

\bibitem{bsharat2023principled}
Sondos~Mahmoud Bsharat, Aidar Myrzakhan, and Zhiqiang Shen.
\newblock 2023.
\newblock Principled instructions are all you need for questioning {LLaMA-1/2}, {GPT-3.5/4}.
\newblock {\em arXiv preprint arXiv:2312.16171}.

\bibitem{wei2022chain}
Jason Wei, Xuezhi Wang, Dale Schuurmans, Maarten Bosma, Fei Xia, Ed~Chi, Quoc~V Le, Denny Zhou, et~al.
\newblock 2022.
\newblock Chain-of-thought prompting elicits reasoning in large language models.
\newblock {\em Advances in Neural Information Processing Systems}, 35:24824--24837.

\bibitem{madaan2022text}
Aman Madaan and Amir Yazdanbakhsh.
\newblock 2022.
\newblock Text and patterns: {F}or effective chain of thought, it takes two to tango.
\newblock {\em arXiv preprint arXiv:2209.07686}.

\bibitem{chu2023survey}
Zheng Chu, Jingchang Chen, Qianglong Chen, Weijiang Yu, Tao He, Haotian Wang, Weihua Peng, Ming Liu, Bing Qin, and Ting Liu.
\newblock 2023.
\newblock A survey of chain of thought reasoning: Advances, frontiers and future.
\newblock {\em arXiv preprint arXiv:2309.15402}.

\bibitem{lin2023generating}
Zhen Lin, Shubhendu Trivedi, and Jimeng Sun.
\newblock 2023.
\newblock Generating with confidence: {U}ncertainty quantification for black-box large language models.
\newblock {\em arXiv preprint arXiv:2305.19187}.

\bibitem{zhou2023navigating}
Kaitlyn Zhou, Dan Jurafsky, and Tatsunori Hashimoto.
\newblock 2023.
\newblock Navigating the grey area: {E}xpressions of overconfidence and uncertainty in language models.
\newblock {\em arXiv preprint arXiv:2302.13439}.

\bibitem{chen2024introducing}
{Chen, Jiefeng and Yoon, Jinsung}.
\newblock 2024.
\newblock Introducing {ASPIRE} for selective prediction in {LLMs}.
\newblock \url{https://blog.research.google/2024/01/introducing-aspire-for-selective.html?m=1}.
\newblock Accessed: 2024-01-20.

\bibitem{paul2023refiner}
Debjit Paul, Mete Ismayilzada, Maxime Peyrard, Beatriz Borges, Antoine Bosselut, Robert West, and Boi Faltings.
\newblock 2023.
\newblock {REFINER}: {R}easoning feedback on intermediate representations.
\newblock {\em arXiv preprint arXiv:2304.01904}.

\bibitem{shinn2023reflexion}
Noah Shinn, Beck Labash, and Ashwin Gopinath.
\newblock 2023.
\newblock Reflexion: {A}n autonomous agent with dynamic memory and self-reflection.
\newblock {\em arXiv preprint arXiv:2303.11366}.

\bibitem{yao2022react}
Shunyu Yao, Jeffrey Zhao, Dian Yu, Nan Du, Izhak Shafran, Karthik Narasimhan, and Yuan Cao.
\newblock 2022.
\newblock {ReAct}: {S}ynergizing reasoning and acting in language models.
\newblock {\em arXiv preprint arXiv:2210.03629}.

\bibitem{xu2023reasoning}
Zhenran Xu, Senbao Shi, Baotian Hu, Jindi Yu, Dongfang Li, Min Zhang, and Yuxiang Wu.
\newblock 2023.
\newblock Towards reasoning in large language models via multi-agent peer review collaboration.
\newblock {\em arXiv preprint arXiv:2311.08152}.

\bibitem{gu2023mamba}
Albert Gu and Tri Dao.
\newblock 2023.
\newblock Mamba: {L}inear-time sequence modeling with selective state spaces.
\newblock {\em arXiv preprint arXiv:2312.00752}.

\end{thebibliography}
